\documentclass[runningheads]{llncs}
\usepackage[T1]{fontenc}
\usepackage{graphicx}
\usepackage{cite}
\usepackage{xcolor}
\usepackage{amsfonts}
\usepackage[hidelinks]{hyperref}
\usepackage{cleveref}
\usepackage{enumitem}
\usepackage{pgfplots}
\usepackage{tikz}
\usetikzlibrary{shapes.geometric, positioning, decorations.pathreplacing, calc, intersections, pgfplots.groupplots}

\usepackage{orcidlink}
\newcommand{\orcidIDs}[1]{\orcidlink{#1}}

\Crefname{section}{Sec.}{Secs.}
\Crefname{equation}{Eq.}{Eqs.}
\Crefname{figure}{Fig.}{Figs.}
\Crefname{table}{Table}{Tables}

\begin{document}

\title{Zero-Shot Brain MRI Inpainting with \\2.5D Unconditional Flow Priors}

\author{Arnela Hadzic\orcidIDs{0009-0007-1476-3729} \and
Franz Thaler\orcidIDs{0000-0002-6589-6560} \and
Simon Johannes Joham\orcidIDs{0009-0009-1817-2904} \and \\
Martin Urschler\orcidIDs{0000-0001-5792-3971}}

\authorrunning{A. Hadzic et al.}

\institute{
Institute for Medical Informatics, Statistics and Documentation, \\Medical University of Graz, Graz, Austria\\
\email{arnela.hadzic@medunigraz.at}}

\maketitle              

\begin{abstract}
Generative inpainting of brain MRI volumes is essential for synthesizing healthy tissue in pathological regions, improving the accuracy and reliability of automated downstream brain analysis applications such as image registration, brain extraction, and segmentation.
However, standard 3D approaches are computationally prohibitive, while efficient 2D slice-wise methods suffer from severe inter-slice discontinuities. Furthermore, traditional models rely on conditional training, requiring task-specific learning of masked inputs.
We propose a zero-shot brain MRI inpainting framework utilizing 2.5D unconditional flow priors to capture spatial context along the superior-inferior axis without the overhead of full 3D convolutions. During training, our flow matching model learns the joint distribution of adjacent axial slice triplets, modeling the manifold of healthy brain anatomy while explicitly excluding pathological regions from the loss function. At inference, the model processes the input triplets autoregressively along the depth axis. We employ the Restora-Flow solver to constrain the unconditional prior using the input mask, achieving accurate zero-shot inpainting.
Evaluations show our 2.5D strategy resolves the structural discontinuities of 2D baselines, synthesizing plausible healthy tissue while maintaining volumetric consistency across the axial, sagittal, and coronal planes.
As a final step, we generate and average an ensemble of multiple stochastic reconstructions to form the final prediction.
Quantitative results benchmarked on the official BraTS 2026 Inpainting Challenge validation set demonstrate the effectiveness of our proposed approach, yielding an SSIM of 0.816 $\pm$ 0.112, MSE of 0.007 $\pm$ 0.005, and PSNR of 22.923 $\pm$ 4.343. Code is available at \url{https://github.com/imigraz/brats2026-inpainting}.

\end{abstract}

\section{Introduction}
Generative inpainting of brain magnetic resonance imaging (MRI) synthesizes plausible healthy tissue in arbitrary regions of the brain. One use case is the synthetic replacement of pathological regions such as brain tumors. This synthesis is essential for standardizing neuroimaging pipelines, as downstream tasks, e.g., image registration, brain extraction, and segmentation, often fail in the presence of severe anatomical anomalies. Recently, diffusion models~\cite{ho2020denoising} and flow matching~\cite{lipman2023flow,liu2023flow} have emerged as the state-of-the-art for modeling complex anatomical distributions~\cite{pinaya2022brain,wolleb2022diffusion,kazerouni2023diffusion,wu2024medsegdiff,hadzic2025flow,ma2025cardiacflow,morehead2026flow}, with flow matching offering faster inference while maintaining comparable or even achieving superior generative performance.

However, generating high-resolution volumetric MRI data remains computationally challenging. While standard 3D architectures provide strong spatial context, memory bottlenecks often necessitate aggressive downsampling~\cite{dorjsembe2022three}, patch-based processing~\cite{bieder2024memory}, or compression into lower-dimensional latent spaces~\cite{zhu2023make}, all of which compromise between preserving fine anatomical details and retaining global context.
Alternatively, 2D slice-wise methods are computationally efficient and preserve in-plane resolution, but processing slices independently causes inter-slice discontinuities. This results in anatomically incoherent reconstructions that fail to preserve continuous anatomical structures across slices. Beyond architectural constraints, traditional inpainting models rely on conditional training. Forcing the network to map masked inputs directly to target images requires task-specific optimization and limits generalizability, as the model overfits to the specific masking distribution seen during training.

We propose a zero-shot framework for brain MRI inpainting that leverages 2.5D unconditional flow priors directly in pixel space, eliminating the need for task-specific, conditional optimization. Developed in the context of the BraTS 2026 Inpainting Challenge~\cite{kofler2023brain}, this approach captures spatial context across axial slices without the computational overhead of full 3D convolutions.
The model learns the manifold of healthy brain anatomy from pathology-affected brain MRI data by explicitly excluding the masked pathological regions when computing gradient updates in the loss function. Importantly, the model is never trained to explicitly predict masked image regions and therefore learns to synthesize anatomical structure independent of the specific mask shape.
To balance computational efficiency with volumetric coherence, we employ an autoregressive strategy at inference time, processing adjacent axial slice triplets to maintain consistency along the superior-inferior axis while simultaneously preserving the original resolution.
Our contributions are summarized as follows:
\begin{itemize}[label=\scriptsize\textbullet]
\item We train an unconditional flow matching model to learn a distribution of healthy brain anatomy, excluding pathological regions from the loss function to ensure anatomical fidelity. 
\item We employ the Restora-Flow~\cite{hadzic2026restora} solver at inference to efficiently perform mask-guided inpainting, enabling accurate synthesis without requiring a conditional training strategy.
\item We introduce a 2.5D approach that utilizes adjacent axial slice triplets to capture joint spatial distributions along the superior-inferior axis, successfully preserving spatial context while remaining computationally efficient.
\item Experimental results demonstrate that our 2.5D strategy resolves inter-slice discontinuities inherent to 2D methods, maintaining anatomical consistency across all imaging planes without compromising in-plane resolution. By aggregating an ensemble of predictions, our approach shows promising performance on the validation dataset of the BraTS 2026 Inpainting Challenge.
\end{itemize}

\section{Methods}
To address the brain MRI inpainting task in a zero-shot manner, we utilize Restora-Flow~\cite{hadzic2026restora}, an efficient mask-guided image restoration framework based on flow matching. Restora-Flow leverages an unconditional flow matching model, pre-trained exclusively on healthy anatomy, as a structural prior. During inference, it applies mask-guidance paired with trajectory correction to reconstruct masked regions without requiring task-specific optimization. To overcome the computational constraints of high-resolution 3D volume generation while preserving cross-slice structural consistency, we introduce a 2.5D formulation. Specifically, the flow matching model is trained to process multi-channel inputs comprising three consecutive axial slices. At inference, we perform an autoregressive propagation along the superior-inferior axis, feeding the inpainted output of preceding slices into the subsequent generation steps to ensure continuous volumetric reconstructions.

\subsection{Flow Matching}
\label{sec:flow_matching}
Flow matching~\cite{lipman2023flow,liu2023flow} defines a generative process by learning a vector field $v_\theta(x_t, t)$ that maps a simple base distribution (e.g., standard Gaussian noise) at $t=0$ to the target data distribution $\mathrm{p}(x)$ at $t=1$ via an ordinary differential equation (ODE):$$\frac{dx}{dt} = v_{\theta}(x_t,t),$$where $\theta$ denotes the learnable parameters of the neural network (e.g., a U-Net~\cite{ronneberger2015u}).

Given a target data sample $x_1 \sim \mathrm{p}(x)$, a noise sample $x_0 \sim \mathcal{N}(0, I)$, and a time step $t \sim \mathcal{U}(0, 1)$, the intermediate noisy sample $x_t$ is constructed as a linear interpolation between the noise and the data: $x_t = (1-t) x_0 + t x_1$. The corresponding target vector field governing this trajectory is given by $u_t(x_t) = x_1 - x_0$. The standard flow matching objective minimizes the mean squared error (MSE) between the predicted and target vector fields~\cite{lipman2023flow}:$$\mathcal{L}_{FM} = \mathbb{E}_{t, x_0, x_1} \left[ \Vert v_\theta(x_t, t) - u_t(x_t) \Vert^2 \right].$$

In our approach, to ensure the model learns strictly from healthy anatomy, we define a binary structural mask $\mathcal{S} = 1 - M_{unhealthy}$, where $M_{unhealthy}$ is the pathology mask provided within the challenge data and apply $\mathcal{S}$ to zero out the unhealthy regions.
The resulting objective $\mathcal{L}_{masked}$ is the spatial mean of the squared vector field differences computed \emph{only} over the healthy pixels:$$\mathcal{L}_{masked} = \mathbb{E}_{t, x_0, x_1} \left[ \frac{\sum \left( \mathcal{S} \odot \Vert v_\theta(x_t, t) - u_t(x_t) \Vert^2 \right)}{\sum \mathcal{S} + \epsilon_{num}} \right],$$where $\odot$ denotes element-wise multiplication along the spatial dimensions, and $\epsilon_{num} = 10^{-8}$ is a small constant added for numerical stability.

During inference, new samples can be generated by starting from noise $x_0 \sim \mathcal{N}(0, I)$ and integrating the ODE. For instance, using the Euler integration scheme, the state is iteratively updated:$$x_{t+\Delta t} = x_t + \Delta_t v_{\theta}(x_t, t).$$

\subsection{2.5D Formulation}
The generation of high-resolution 3D medical volumes imposes severe memory and computational bottlenecks. While pure 2D slice-wise approaches are efficient to train, they ignore spatial dependencies along the respective third axis, leading to cross-slice structural inconsistencies when aggregated into a volume. Conversely, fully 3D architectures preserve this continuity but are heavily constrained by memory requirements and extended training times, especially in generative modeling, whereas latent space methods mitigate these memory issues at the expense of smoothing over high-frequency structural details.

To address this trade-off, we introduce a 2.5D formulation. Specifically, we configure the unconditional flow matching model (see \Cref{sec:flow_matching}) to process multi-channel inputs comprising three consecutive axial slices. Mathematically, for a target slice at depth $d$, the input tensor is constructed as $X \in \mathbb{R}^{3 \times H \times W}$, where the channel dimension ($C = 3$) represents the localized sequence of adjacent axial slices $\{d-1, d, d+1\}$. Through this formulation, the network inherently learns the joint probability distribution of the localized 3D neighborhood, and unconditionally generates a structurally consistent stack of three adjacent slices. Integrating this local spatial context directly into the generative prior yields anatomically continuous reconstructions that outperform standard 2D baselines (see \Cref{sec:2D_vs_2_5D}), while not imposing limitations of a full 3D model.

\subsection{Inpainting with Restora-Flow}
To reconstruct the masked regions (voids) while preserving the intact anatomy, we employ Restora-Flow~\cite{hadzic2026restora} at inference. Unlike standard unconditional generation, Restora-Flow modifies the ODE integration steps to incorporate mask-guidance.

Given a voided input image $y$ and its corresponding binary mask $M$ (where $M=1$ denotes the known, intact tissue), the inference process initializes with standard Gaussian noise. At each integration step, the predicted vector field is explicitly conditioned to combine the generated content within the masked region with the intact tissue outside the mask:
$$ x_{t}' = M \odot y' + (1-M) \odot x_{t} , $$
where $y'$ is a noisy interpolation of the input $y$:
$$ y' = t y + (1-t) \epsilon, \quad  \epsilon \sim \mathcal{N}(0,I) .$$
Restora-Flow additionally applies trajectory correction steps during the solver schedule to iteratively refine the boundary transitions between the known and reconstructed tissues. The correction involves extrapolating the sample toward the data manifold, and re-injecting noise to bring the sample back into the generative process. We adopt this correction directly, referring readers to~\cite{hadzic2026restora} for the mathematical derivations.

To apply the 2.5D flow-based restoration to a full 3D volume, we perform an autoregressive propagation along the superior-inferior axis. Let $V$ represent the complete masked volume. We iteratively process the volume slice-by-slice and consider only those slices that need to be inpainted. For a given target slice at index $d$, we extract a 3-slice input window consisting of $\{d-1, d, d+1\}$. 
Note that if the mask includes the very first or last slice of the volume, making $d-1$ respectively $d+1$ undefined, we simply skip processing for that slice, though alternative boundary handling for example by replicating the border slice could also be employed.
In order to maintain cross-slice structural consistency, the preceding slice $d-1$ is not taken from the original voided volume; it is instead the inpainted output from the previous generation step. 
By integrating the output of slice $d-1$ into the input tensor for slice $d$, the current reconstruction is autoregressively conditioned on the anatomy synthesized in the previous step. 
This iterative conditioning process ensures that the generated anatomical structures propagate smoothly through the entire axial stack of the masked region, effectively mitigating the stacking artifacts characteristic of independent 2D slice-wise generation (see \Cref{sec:2D_vs_2_5D} and \Cref{fig:2D_vs_2_5D}).

\section{Experimental Setup}
\paragraph{\textbf{Dataset.}}
We utilize the BraTS 2026 Inpainting Challenge dataset~\cite{kofler2023brain,karargyris2023federated}, comprising 1,251 T1-weighted brain MRI scans. 
The original volumes have spatial dimensions of $240 \times 240 \times 155$ with an isotropic voxel spacing of 1 mm\textsuperscript{3}. For each training scan, the dataset provides five items: the ground truth T1-weighted image (\texttt{t1n}), the masked input image (\texttt{t1n-voided}), the combined inpainting region (\texttt{mask}), as well as isolated masks for the actual tumor (\texttt{mask\_unhealthy}) and synthetically voided healthy tissue (\texttt{mask\_healthy}).
During the development phase, we perform a patient-wise split of the provided dataset, allocating 90\% of the cases for training and 10\% for internal validation. All images are centrally cropped to $208 \times 208 \times 144$ and their intensity values are normalized to the range [-1, 1] using the minimum and maximum intensity value of each respective input image.
Following the generation of the inpainted regions, the predicted intensity values are mapped back to their original scale, and the resulting volumes are padded back to their original dimensions and spatial origin. 
The official validation as performed through the submission system of the BraTS 2026 Inpainting Challenge was conducted on 219 validation cases, for which ground truth images were withheld from participants.
For the final challenge submission, we retrained the model on the complete training dataset using the optimal hyperparameters identified during internal validation.

\paragraph{\textbf{Implementation Details.}}
The unconditional flow matching models were trained using the AdamW~\cite{loshchilov2019adamw} optimizer with a learning rate of $1 \times 10^{-4}$ and a batch size of 16. The vector field was parameterized using a 2D U-Net, adapted to process adjacent axial slice triplets by using three input and three output channels. 
The model used for internal validation was trained for 2,000 epochs, whereas the final model retrained on the complete training dataset was optimized to 10,000 epochs. 
Training this final model required approximately 100 hours on a single NVIDIA A100 GPU with 40 GB memory. 
During inference, we employed an ensemble strategy to aggregate multiple stochastic predictions. Individual reconstructions were generated using the Euler solver with 32 ODE steps (see \Cref{sec:ablation}). 
Additionally, the number of trajectory correction steps for the Restora-Flow sampling was set to 1, as recommended in~\cite{hadzic2026restora}.
The average inference time to reconstruct a single 3D volume using 32 ODE steps took 3 minutes and 48 seconds for an average of 97 masked slices.
Inference for an ensemble with $N=10$ and parallel generation lasted roughly 24 minutes, which is around 36\% faster than sequential processing.
Inference times were measured on a single NVIDIA GeForce RTX 3090 Ti GPU with 24 GB memory.

\paragraph{\textbf{Metrics.}}
Since there is no ground truth for pathological regions, the evaluation is restricted to the synthetically voided healthy tissue. The quality of the inpainted healthy regions is assessed via Structural Similarity Index Measure (SSIM), Peak Signal-to-Noise Ratio (PSNR), and Mean Squared Error (MSE).

\section{Results}
\subsection{Ablation on ODE Integration Steps}
\label{sec:ablation}
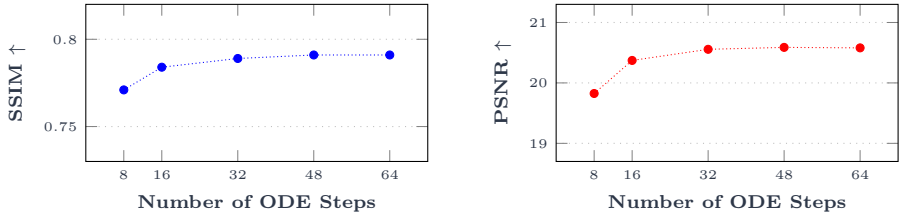
\begin{figure*}[h]
    \begin{minipage}{.5\textwidth}
        \centering
        \definecolor{charcoal}{HTML}{2D3748}
\definecolor{bordergray}{HTML}{E2E8F0}
\pgfplotsset{compat=1.18}
\begin{tikzpicture}
    \begin{axis}[
        width=\linewidth,
        height=0.6\linewidth,
        xlabel={\textbf{Number of ODE Steps}},
        ylabel={\textbf{SSIM $\uparrow$}},
        xlabel style={text=charcoal,font=\scriptsize},
        ylabel style={text=charcoal,font=\scriptsize,yshift=0.5pt},
        xmin=0, xmax=72,
        ymin=0.73, ymax=0.82,
        xtick={8, 16, 32, 48, 64},
        ytick={0.75, 0.8},
        tick label style={font=\tiny\sffamily, text=charcoal},
        ymajorgrids=true,
        grid style=dotted,
        axis background/.style={fill=white}
    ]

    % 3. Arithmetic Mean
    \addplot[
        color=blue, 
        densely dotted, 
        mark=*, 
        mark options={solid, scale=0.75}
    ] coordinates {
        (8, 0.771) 
        (16, 0.784)
        (32, 0.789) 
        (48, 0.791) 
        (64, 0.791)
    };

    \end{axis}
\end{tikzpicture}
    \end{minipage}
    \hspace{2pt}
    \begin{minipage}{.5\textwidth}
        \centering
        \definecolor{charcoal}{HTML}{2D3748}
\definecolor{bordergray}{HTML}{E2E8F0}
\pgfplotsset{compat=1.18}
\begin{tikzpicture}
    \begin{axis}[
        width=\linewidth,
        height=0.6\linewidth,
        xlabel={\textbf{Number of ODE Steps}},
        ylabel={\textbf{PSNR $\uparrow$}},
        xlabel style={text=charcoal,font=\scriptsize},
        ylabel style={text=charcoal,font=\scriptsize,yshift=0.5pt},
        xmin=0, xmax=72,
        ymin=18.7, ymax=21.3,
        xtick={8, 16, 32, 48, 64},
        ytick={19, 20, 21},
        tick label style={font=\tiny\sffamily, text=charcoal},
        ymajorgrids=true,
        grid style=dotted,
        axis background/.style={fill=white}
    ]

    % 3. Arithmetic Mean
    \addplot[
        color=red, 
        densely dotted, 
        mark=*, 
        mark options={solid, scale=0.75}
    ] coordinates {
        (8, 19.825) 
        (16, 20.372)
        (32, 20.556) 
        (48, 20.588) 
        (64, 20.580)
    };

    \end{axis}
\end{tikzpicture}
    \end{minipage}
    \caption{
    Impact of the number of ODE steps on reconstruction quality, measured via SSIM (left) and PSNR (right) on the internal validation set using a single reconstruction per image obtained from a model trained for 2,000 epochs. 
    }
    \label{fig:ablation_ode}
\end{figure*}

We first investigate the impact of the number of ODE integration steps on model performance using the internal validation set. As illustrated in \Cref{fig:ablation_ode}, increasing the number of inference steps generally improves reconstruction quality, but at a higher computational cost. 
We observe that the quantitative metrics stabilize after 32 ODE steps, representing the optimal configuration for efficient sampling.

% ODE steps ablation plot

\begin{figure*}[t]
    \centering
    \resizebox{\linewidth}{!}{%
    \input{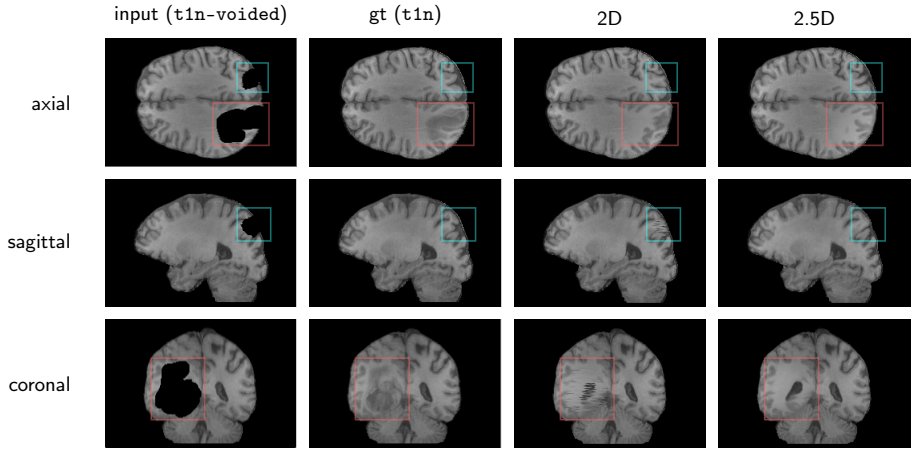}
    }
    \caption{Qualitative comparison of volumetric reconstructions between the 2D baseline and the proposed 2.5D strategy. Rows denote the axial, sagittal, and coronal anatomical planes. Columns from left to right display the masked input (\texttt{t1n-voided}), the ground truth (\texttt{t1n}), the 2D reconstruction, and the 2.5D reconstruction. The boxes indicate the unhealthy (red) and healthy (cyan) masked region. While the 2D approach introduces severe inter-slice discontinuities visible in the sagittal and coronal planes, the 2.5D strategy successfully preserves spatial consistency across all planes. The reconstructions are single-pass predictions of models trained for 2,000 epochs using 32 ODE steps.
    }
    \label{fig:2D_vs_2_5D}
\end{figure*}

\subsection{Comparison of 2D and 2.5D Strategies}
\label{sec:2D_vs_2_5D}
Training an unconditional flow matching model on single 2D axial slices and applying Restora-Flow to individually reconstruct the full 3D volume in a slice-by-slice manner introduces severe spatial inconsistencies in the superior-inferior axis as observable in the sagittal and coronal planes (see \Cref{fig:2D_vs_2_5D}). 

To address this limitation, our proposed 2.5D autoregressive strategy explicitly incorporates spatial context through processing adjacent axial slice triplets. As visualized in \Cref{fig:2D_vs_2_5D}, this approach successfully resolves inter-slice discontinuities, which is also reflected in the quantitative scores reported in \Cref{tab:2D_vs_2_5D}, where the 2.5D strategy outperforms the standard 2D baseline across all metrics. For a fair comparison, both approaches were evaluated using model checkpoints at 2,000 epochs with 32 ODE integration steps on the internal validation set.

\begin{table}[h]
    \centering
    \caption{Quantitative comparison of reconstruction quality between the standard 2D and the proposed 2.5D strategies on the internal validation set. Scores were computed from a single reconstruction using models trained for 2,000 epochs and 32 ODE steps.}
    \label{tab:2D_vs_2_5D}
    \setlength{\tabcolsep}{14pt}
    \begin{tabular}{lccc}
        \hline
        Strategy &  SSIM $\uparrow$ & MSE $\downarrow$ & PSNR $\uparrow$ \\
        \hline
        2D   & 0.727 $\pm$ 0.153 & 0.014 $\pm$ 0.010 & 18.899 $\pm$ 2.804 \\
        2.5D & 0.789 $\pm$ 0.131 & 0.010 $\pm$ 0.008 & 20.556 $\pm$ 3.004 \\
        \hline
    \end{tabular}
\end{table}

\subsection{Generative Ensemble}
\label{sec:ensemble}
\begin{figure*}[t]
    \centering
    \resizebox{\linewidth}{!}{%
    \definecolor{charcoal}{HTML}{2D3748}
\definecolor{bordergray}{HTML}{E2E8F0}

\definecolor{ode32yellow}{RGB}{236, 203, 74} 
\definecolor{ode16red}{RGB}{209, 65, 62}    
\definecolor{ode8blue}{RGB}{68, 127, 186}  

\pgfplotsset{
    commonaxis/.style={
        % Width and height adjusted to fit nicely side-by-side on a page
        width=5.8cm,
        height=4cm,
        xlabel={\textbf{Number of Reconstructions ($N$)}},
        xlabel style={text=charcoal,font=\scriptsize},
        ylabel style={text=charcoal,font=\scriptsize,yshift=-13pt},
        xmin=0, xmax=11,
        xtick={1, 2, 4, 6, 8, 10},
        tick label style={font=\tiny\sffamily, text=charcoal},
        ymajorgrids=true,
        grid style=dotted,
        axis background/.style={fill=white},
        % Legend shared configuration
        legend style={
            at={(0.5,-0.45)}, 
            anchor=north, 
            legend columns=-1, 
            font=\scriptsize\sffamily, 
            text=charcoal,
            draw=bordergray,
            /tikz/every even column/.append style={column sep=0.5cm}
        }
    },
    % Common plot line style to match your snippet
    myplot/.style={
        thick,
        mark=*,
        mark options={solid, scale=0.75}
    }
}

\begin{tikzpicture}
\begin{groupplot}[
    group style={
        group name=ablationplots,
        rows=1, columns=3,
        horizontal sep=1.8cm,
    },
    commonaxis
]

% ========================================================
% Plot 1: SSIM 
% ========================================================
\nextgroupplot[
    ylabel={\textbf{SSIM $\uparrow$}},
    ymin=0.78, ymax=0.85,
    ytick={0.8, 0.83},
]
    \addplot[color=ode8blue, myplot] coordinates {(1, 0.788) (2, 0.802) (4, 0.812) (6, 0.817) (8, 0.818) (10, 0.819)};

    \addplot[color=ode16red, myplot] coordinates {(1, 0.802) (2, 0.815) (4, 0.822) (6, 0.827) (8, 0.828) (10, 0.829)};

    \addplot[color=ode32yellow, myplot] coordinates {(1, 0.808) (2, 0.819) (4, 0.825) (6, 0.827) (8, 0.829) (10, 0.829)};

    % Optional asymptote line
    \addplot[color=bordergray, thick, densely dotted, mark=none, domain=0:35, samples=2] {0.99};

% % ========================================================
% % Plot 2: RMSE 
% % ========================================================
% \nextgroupplot[
%     ylabel={\textbf{RMSE $\downarrow$}},
%     scaled y ticks=false,
%     yticklabel style={
%         /pgf/number format/fixed,
%         /pgf/number format/precision=3
%     },
%     ymin=0.075, ymax=0.105,
%     ytick={0.08, 0.09, 0.1},
% ]
%     % ODE=8
%     \addplot[color=ode8blue, myplot] coordinates {(1, 0.1005) (2, 0.0949) (4, 0.0908) (6, 0.0887) (8, 0.0881) (10, 0.0877)};
%     \addlegendentry{ODE=8}
    
%     % ODE=16
%     \addplot[color=ode16red, myplot] coordinates {(1, 0.0923) (2, 0.0872) (4, 0.0851) (6, 0.0827) (8, 0.0821) (10, 0.0820)};
%     \addlegendentry{ODE=16}

%     % ODE=32
%     \addplot[color=ode32yellow, myplot] coordinates {(1, 0.0902) (2, 0.0852) (4, 0.0837) (6, 0.0826) (8, 0.0822) (10, 0.0817)};
%     \addlegendentry{ODE=32}

% ========================================================
% Plot 2: MSE 
% ========================================================
\nextgroupplot[
    ylabel={\textbf{MSE $\downarrow$}},
    scaled y ticks=false,
    yticklabel style={
        /pgf/number format/fixed,
        /pgf/number format/precision=3
    },
    ymin=0.0065, ymax=0.013,
    ytick={0.008, 0.010, 0.012},
]
    % ODE=8
    \addplot[color=ode8blue, myplot] coordinates {(1, 0.0116) (2, 0.0103) (4, 0.0095) (6, 0.0091) (8, 0.0089) (10, 0.0089)};
    \addlegendentry{ODE=8}
    
    % ODE=16
    \addplot[color=ode16red, myplot] coordinates {(1, 0.0099) (2, 0.0088) (4, 0.0083) (6, 0.0078) (8, 0.0078) (10, 0.0077)};
    \addlegendentry{ODE=16}

    % ODE=32
    \addplot[color=ode32yellow, myplot] coordinates {(1, 0.0096) (2, 0.0085) (4, 0.0081) (6, 0.0079) (8, 0.0079) (10, 0.0078)};
    \addlegendentry{ODE=32}

% ========================================================
% Plot 3: PSNR
% ========================================================
\nextgroupplot[
    ylabel={\textbf{PSNR $\uparrow$}},
    ymin=19, ymax=22,
    ytick={20, 21},
]
    \addplot[color=ode8blue, myplot] coordinates {(1, 19.561) (2, 20.064) (4, 20.470) (6, 20.663) (8, 20.721) (10, 20.745)};

    \addplot[color=ode16red, myplot] coordinates {(1, 20.307) (2, 20.776) (4, 20.960) (6, 21.203) (8, 21.297) (10, 21.317)};

    \addplot[color=ode32yellow, myplot] coordinates {(1, 20.579) (2, 21.022) (4, 21.193) (6, 21.280) (8, 21.349) (10, 21.403)};

    % Optional asymptote line 
    \addplot[color=bordergray, thick, densely dotted, mark=none, domain=0:35, samples=2] {34.5};

\end{groupplot}
\end{tikzpicture}
    }
    \caption{Impact of ensemble size ($N$) and ODE steps on reconstruction quality, measured via SSIM (left), MSE (center), and PSNR (right) on a subset of 20 cases from the internal validation set. Scores were obtained using a model trained for 2,000 epochs.
    }
    \label{fig:ensemble_ablation}
\end{figure*}
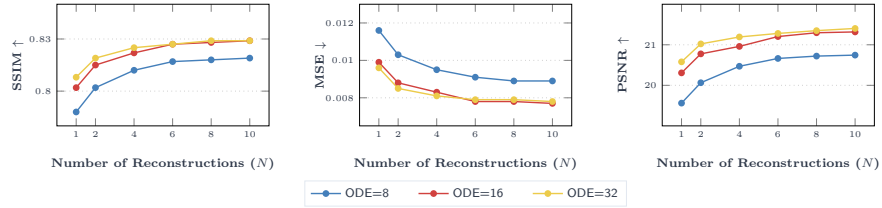

Variations in the initial noise sample $x_0$ (see \Cref{sec:flow_matching}) allow the model to explore different trajectories on the learned manifold, generating distinct but valid anatomical reconstructions. As shown in \Cref{fig:ensemble_ablation}, we observe that aggregating multiple predictions via voxel-wise averaging improves quantitative results. 
Increasing the ensemble size $N$ consistently improves performance, with SSIM and PSNR increasing as MSE decreases. Performance gains are most evident when moving from a single reconstruction ($N=1$) to an ensemble size of $N=6$, with improvements plateauing as the ensemble size approaches $N=10$. While higher integration steps (e.g., ODE=32) establish a superior baseline, the ensemble effect remains consistent across all solver configurations. Therefore, we selected $N=10$ and 32 ODE steps for the final submission.

\subsection{Evaluation on the Challenge-Provided Validation Set}
Table~\ref{tab:validation} summarizes the quantitative results on the challenge validation set, comparing single-sample reconstructions against our ensemble strategy. In line with the findings from our ablation study on the internal validation set (see \Cref{fig:ensemble_ablation}), generative ensembling yields consistent quantitative improvements. Increasing the ensemble size from $N=1$ to $N=10$ progressively improves SSIM, PSNR, and MSE metrics, while slightly reducing the standard deviation for SSIM and MSE, suggesting improved model stability across the validation set.

\begin{table}[h]
    \centering
    \caption{Quantitative results on the challenge validation set. We compare single reconstructions ($N=1$) against ensembles ($N=5, 10$) obtained via voxel-wise averaging. We used 32 ODE steps per prediction and a model trained for 10,000 epochs.
    }
    \label{tab:validation}
    \setlength{\tabcolsep}{14pt}
    \begin{tabular}{lccc}
        \hline
        Model &  SSIM $\uparrow$ & MSE $\downarrow$ & PSNR $\uparrow$ \\
        \hline
        $N=1$   & 0.793 $\pm$ 0.122 & 0.009 $\pm$ 0.007 & 21.888 $\pm$ 4.328 \\
        $N=5$   & 0.813 $\pm$ 0.113 & 0.007 $\pm$ 0.006 & 22.793 $\pm$ 4.335 \\
        $N=10$  & 0.816 $\pm$ 0.112 & 0.007 $\pm$ 0.005 & 22.923 $\pm$ 4.343 \\
        \hline
    \end{tabular}
\end{table}

\begin{figure}[p]
    \centering
    \resizebox{\linewidth}{!}{%
    \input{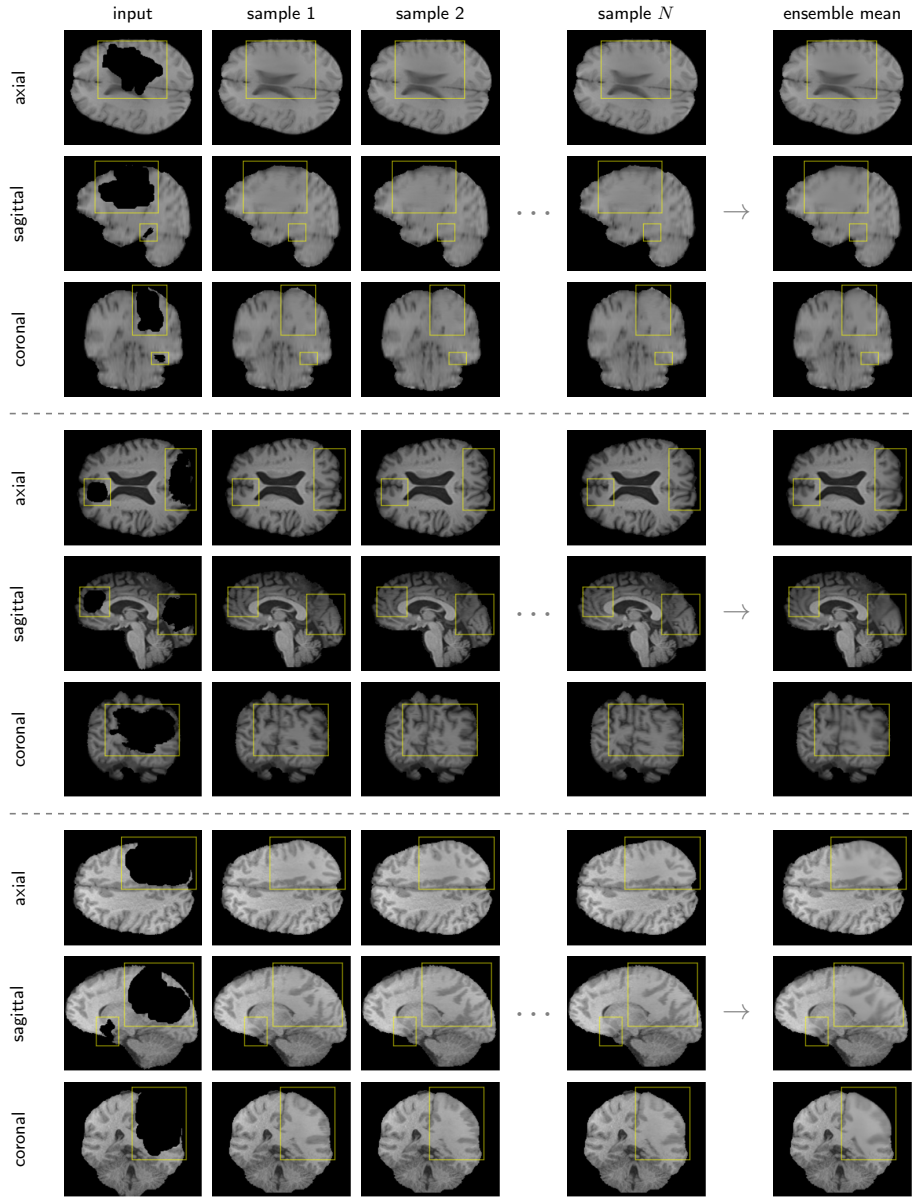}
    }
    \caption{
    Qualitative inpainting results of the proposed 2.5D approach on the official validation set where the ground truth is unavailable.
    The figure highlights three distinct patient cases for which we show the axial, sagittal, and coronal planes for the voided input (col~1), three distinct single reconstructions (col~2-4), and the final ensemble mean with $N=10$ (col~5).
    The single reconstructions were obtained from a model trained for 10,000 epochs using 32 ODE steps.
    Yellow boxes indicate the masked healthy and unhealthy region for which a distinction is not possible.
    The results demonstrate the effectiveness of our approach, producing anatomically plausible reconstructions that maintain structural consistency across all three volumetric planes.
    }
    \label{fig:validation}
\end{figure}

As shown in \Cref{fig:validation}, exemplary single predictions (\texttt{sample~1}, \texttt{sample~2} and \texttt{sample~N}) yield high-quality reconstructions.
The model generates healthy tissue that maintains anatomical continuity with surrounding structures, however, it also allows for some anatomical variability especially for regions distant to the mask's border (see, e.g., row~7, col~2-4).
Compared to single predictions, the ensemble strategy improves quantitative metrics (see \Cref{tab:validation}) but introduces a trade-off in visual fidelity.
Because voxel-wise averaging of independent stochastic predictions acts as a low-pass filter, it minimizes voxel-wise error, benefiting all metrics at the expense of high-frequency anatomical textures.
Consequently, ensemble outputs appear more smoothed than single-pass predictions as can be seen in \Cref{fig:validation} (\texttt{ensemble mean}).
Therefore, while generative ensembling provides the highest quantitative performance, a single reconstruction may be preferable when sharp, highly detailed visual fidelity is the primary objective.

To assess the versatility of our method, we selected three diverse exemplary cases from the validation set. 
In general, the reconstructed image regions appear realistic and maintain structural continuity (see also \Cref{fig:2D_vs_2_5D}). 
However, visual fidelity naturally correlates with mask size and anatomical complexity. 
For example, closer inspection of the axial and coronal views shows that reconstructed regions for larger masks exhibit slightly larger homogeneous areas as compared to the unperturbed other cerebral hemisphere of the same image (\Cref{fig:validation}, rows 3 and 7). 
Additionally, complex features like the ventricles can occasionally remain incompletely recovered, as shown in row 9 of \Cref{fig:validation}.
Overall, our 2.5D method demonstrates effectiveness for inpainting brain MRI data that is not only consistent in 3D but also visually aligned with the specific visual characteristics of the respective image.

\section{Conclusion}

We proposed a framework for zero-shot brain MRI inpainting based on 2.5D unconditional flow priors which was trained on adjacent axial slice triplets of healthy brain anatomy.
Further, we employ the Restora-Flow solver at inference to perform efficient mask-guided inpainting, while constraining the unconditional prior with an input mask.
By processing axial slice triplets, we showed that the proposed 2.5D method successfully preserves spatial context across axial slices and resolves the structural discontinuities of 2D baselines.
In addition, we assessed an ensemble strategy by aggregating multiple stochastic predictions, which improves quantitative metrics but introduces a trade-off in visual quality.
Our method for inpainting healthy brain MRI data is effective in synthesizing spatially consistent 3D regions for arbitrary masks with high visual fidelity.

%\section*{Acknowledgements}
%This research was funded in whole or in part by the Austrian Science Fund (FWF) 10.55776/PAT1748423. The computational results have been achieved using the Austrian Scientific Computing (ASC) infrastructure.

\clearpage
\bibliographystyle{splncs04}
\bibliography{references}

\end{document}